\pdfoutput=1

\documentclass[11pt]{article}

\usepackage{acl}

\usepackage{times}
\usepackage{latexsym}

\usepackage[T1]{fontenc}

\usepackage[utf8]{inputenc} 
\usepackage[T1]{fontenc}    
\usepackage{hyperref}       
\usepackage{url}            
\usepackage{booktabs}       
\usepackage{amsfonts}       
\usepackage{nicefrac}       
\usepackage{microtype}      
\usepackage{xcolor}         
\usepackage{amsmath}
\usepackage{amssymb}
\usepackage{multirow}
\usepackage[ruled]{algorithm2e}
\usepackage{float}
\usepackage{ulem}
\usepackage{microtype}
\usepackage{inconsolata}
\usepackage{graphicx}
\usepackage{subfigure}
\usepackage{makecell}
\allowdisplaybreaks[4]

\title{Improving Factual Consistency of News Summarization by Contrastive Preference Optimization}

\author{Huawen Feng$^{1,2}$\thanks{\ \ This work was conducted when Huawen Feng was interning at Alibaba.}, Yan Fan$^{2}$, Xiong Liu$^{2}$, Ting-En Lin$^{2}$, Zekun Yao$^{1}$, \\ 
\textbf{Yuchuan Wu$^{2}$, Fei Huang$^{2}$, Yongbin Li$^{2\dagger}$, Qianli Ma$^{1}$\thanks{\textsuperscript{\textdagger} Qianli Ma and Yongbin Li are corresponding authors.}}  \\
  $^{1}$School of Computer Science and Engineering, South China University of Technology, China\\
  $^{2}$Alibaba Group\\
  \texttt{541119578@qq.com}, \texttt{99722503@qq.com}, \texttt{qianlima@scut.edu.cn} \\ 
  \texttt{\{fanyan.fy,peter.lx,ting-en.lte,shengxiu.wyc,f.huang,shuide.lyb\}@alibaba-inc.com}
}

\begin{document}
\maketitle
\begin{abstract}
Despite the recent progress in news summarization made by large language models (LLMs), they often generate summaries that are factually inconsistent with original articles, known as "hallucinations" in text generation. Unlike previous small models (e.g., BART, T5), current LLMs make fewer silly mistakes but more sophisticated ones, such as imposing cause and effect, adding false details, overgeneralizing, etc. These hallucinations are challenging to detect through traditional methods, which poses great challenges for improving the factual consistency of text summarization. In this paper, we propose \textbf{C}ontrastive \textbf{P}reference \textbf{O}ptimization (\textbf{CPO}) to disentangle the LLMs' propensities to generate faithful and fake content. Furthermore, we adopt a probing-based specific training method to improve their capacity of distinguishing two types of propensities. In this way, LLMs can execute the instructions more accurately and have enhanced perception of hallucinations. Experimental results show that CPO significantly improves the reliability of summarization based on LLMs. Our code and data are available at \url{https://github.com/201736621051/CPO}.
\end{abstract}

\section{Introduction}
\label{sec:Introduction}
\begin{figure}[t!]
	\centering
	\includegraphics[width=0.9\linewidth]{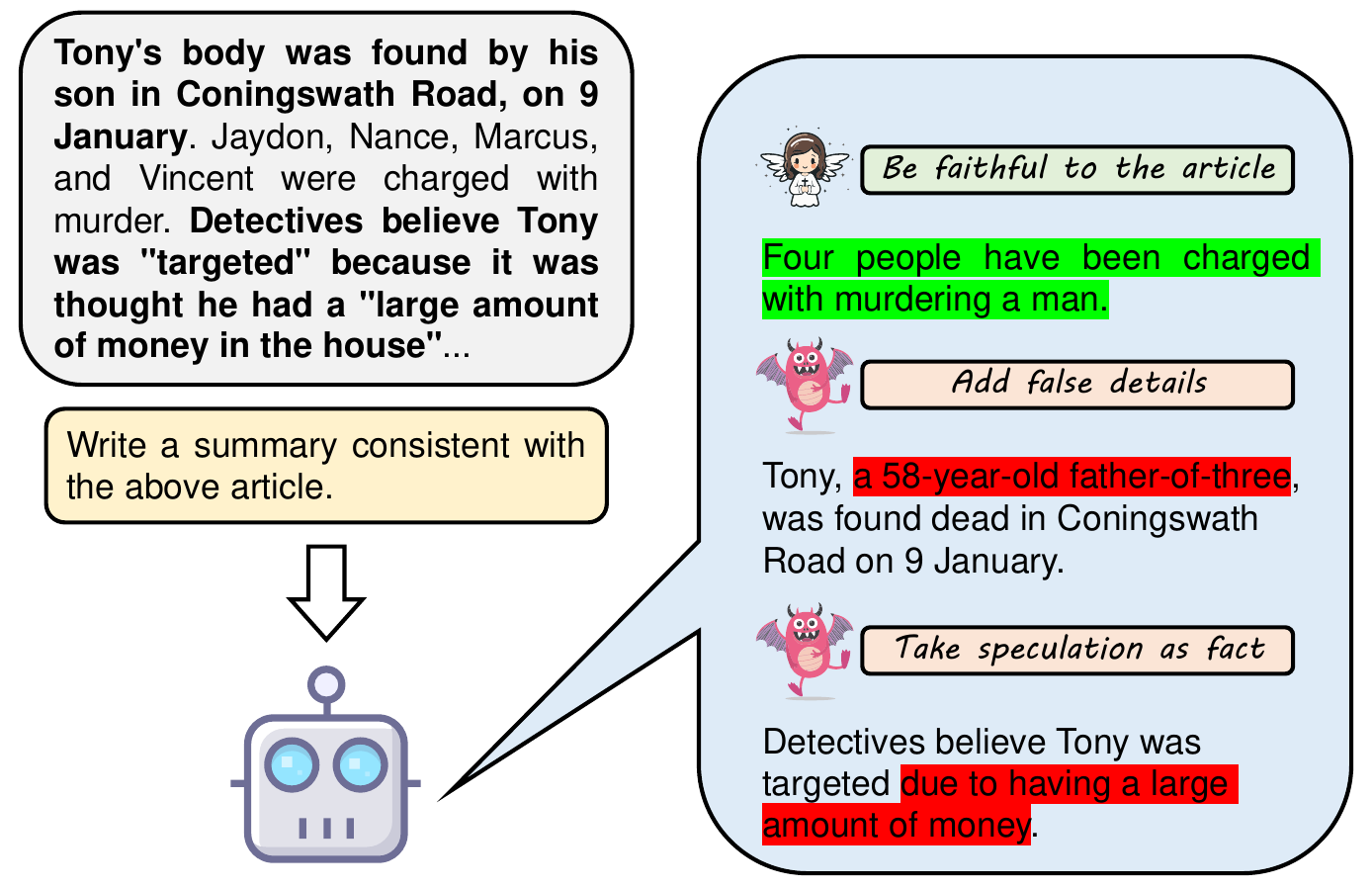}
    \caption{The diagram of the LLMs' propensities to generate faithful and fake content. In abstractive summarization, the model is supposed to generate a factually consistent summary with the preference to be faithful to the context. However, it often hallucinates with the preference to over-imagine with internal knowledge.}
    \label{Fig:intro}
\end{figure}

Although recent pre-trained language models have significantly boosted the performance of abstractive summarization~\cite{text2019liu, bart, t5}, the hallucination problem - that models generate summaries that are factually inconsistent with the source text - remains difficult to resolve. As Figure~\ref{Fig:intro} shows, we expect the model to generate reliable summaries (understand the source text and only generate the faithful content). Still, it often hallucinates and over-imagines, which means the model outputs fake content without supporting evidence in the original article. Worse still, the generation usually involves both types of content, making the summaries so full of half-truths that the hallucinations are much more hidden.

Early methods for improving factual consistency use post-processing models~\cite{correct}, which correct summaries with hallucinations, but they rely on external resources to obtain the error correction capability. ~\citet{feedback} introduces human revisions to achieve better performance, but data collection is still difficult and costly. Besides, these two-stage methods have a complicated structure, consisting of summary generation and correction models. Considering that, some studies try to solve hallucinations holistically during the pre-training stage~\cite{pegasus2020zhang, factpegasus}. They design a new pre-training objective with sentence selection strategies, encouraging the model to generate a faithful summary. However, pre-training requires enormous computational resources, especially for large language models (LLMs).

Moreover, some methods adopt contrastive learning~\cite{cliff} in fine-tuning to teach the model to distinguish between true and false more clearly. To construct negative samples, they modify the references by entity swapping and masking-and-filling. Unfortunately, these auto-generated negative samples are inconsistent with the distribution of errors made by LLMs in real scenarios. \citet{Benchmarking} point that instruction-tuned models have much stronger summarization abilities than previous fine-tuned ones. Current LLMs make fewer silly mistakes (e.g., entity confusion, irrelevant information generation) but more sophisticated ones~\cite{pu2023summarization}. For example, they fill in the details related to but not directly supported by the source text. Sometimes, they rewrite original sentences by imposing cause and effect or taking speculation as fact. These mistakes are difficult to mimic by traditional perturbation-based approaches~\cite{TrueTeacher}.

\begin{figure}[t!]
	\centering
	\includegraphics[width=1.0\linewidth]{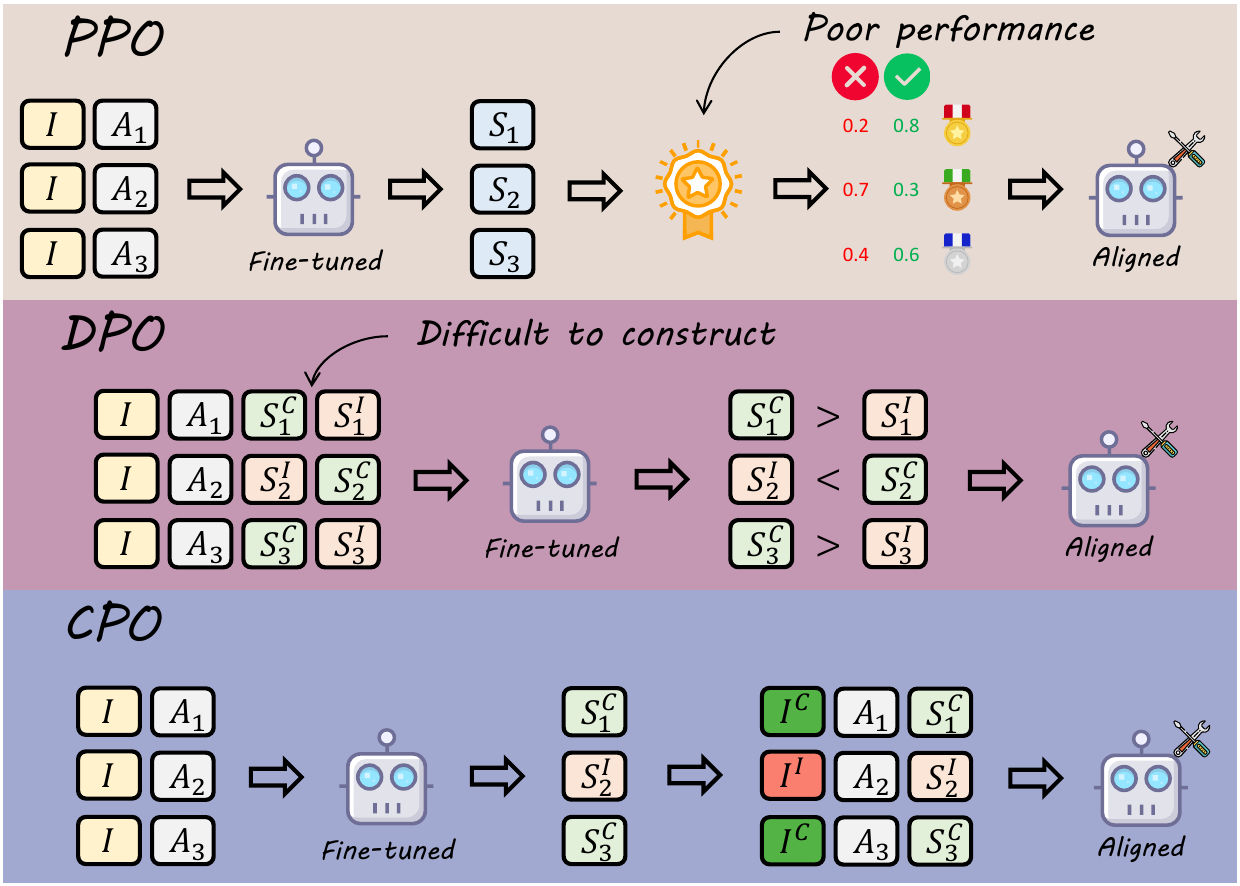}
    \caption{The diagram of our approach compared with methods based on reinforcement learning.}
    \label{Fig:rlhf}
\end{figure}

With the rapid development of LLMs, designing prompts based on the chain of thoughts (COT)~\cite{zhao2023verifyandedit, Element-aware} attracts scholarly attention. The models are posed with several questions about the critical content in the source text before final summarization, serving as contextual clues to guide models to generate factually consistent summaries. Nevertheless, these methods are sensitive to the domain because they do not fundamentally improve the LLMs' reliability. Inspired by preference optimization, many methods use reinforcement learning~\cite{factuallyconsistent, zablotskaia2023calibrating} with entailment feedback (RLEF) to ameliorate hallucination problems. As Figure~\ref{Fig:rlhf} shows, PPO-based methods~\cite{PPO} (Proximal policy optimization) train a Natural Language Inference (NLI) model for consistency detection and then regard it as the reward model in reinforcement learning. However, it is challenging for hallucinations generated by LLMs to be detected through traditional NLI methods. Therefore, the performance of these reward models constrains the training of summarization models. On the other hand, DPO-based methods~\cite{DPO, PPOPAIR} require paired data with preference annotation, which is difficult to construct. Otherwise, reinforcement learning is usually unstable, and rewards are easily over-optimized~\cite{PPOUN}.

The problems mentioned above motivated us to propose \textbf{C}ontrastive
\textbf{P}reference \textbf{O}ptimization (\textbf{CPO}) which disentangles the LLMs' propensities to generate faithful and fake content. Furthermore, we dynamically probe for the model's distinguishing capacity for consistency and inconsistency and train the target layers in an SFT mode, getting rid of the reliance on Reinforcement Learning (RL) frameworks.

This work makes three main contributions:
\begin{itemize}
\item We point out the problem of applying previous methods to summarization based on LLMs through a detailed analysis.
\item We construct a new summarization dataset for training LLMs - \textbf{LESSON}\footnote{The dataset will be released soon.} - \textbf{L}arg\textbf{E} language models' \textbf{S}ummaries with \textbf{S}entence-level c\textbf{ON}sistency annotation. 
\item We propose \textbf{CPO} with probing-based specific training, which can be directly employed in an SFT mode, significantly improving factual consistency without complex data annotation and format requirements.
\end{itemize}

\section{Related Work}

\subsection{Evaluating Factual Consistency}
The problem of hallucinations is inevitable in text summarization, so how to evaluate the factual consistency is a crucial technique. It can be used to measure the summarization reliability and even construct a new summarization dataset~\cite{eval2020, summac}. Inspired by NLI and QA, some methods employ them to assess the summaries~\cite{FEQA, Joshua, Asking}. However, these traditional methods do not work well in LLMs' summaries, for they can hardly detect the subtler mistakes hidden in a longer text. Consequently, the evaluation metrics limit the quality of the constructed summarization dataset. Benefiting from the development of LLMs, ChatGPT and GPT-4 can provide a very accurate assessment~\cite{evalbyllm, Human-like, AreLLMs}, but how to design an appropriate prompt suitable for the domain requires more effort.

\subsection{Probing for Truthfulness}
\label{sec:Probing}
Recent works~\cite{pipeline, KnowledgeNeurons, EditGPT, DoLa} suggest that language models contain latent and interpretable structures related to factuality. Meanwhile, some studies also try to understand the cause of hallucinations~\cite{KnowWhatTheyKnow, Self-critiquing, LatentKnowledge}. Through the hidden states or activation space, these studies observe whether the model can distinguish true output from false one~\cite{EmergentWorld, Relativerepresentations}. An interesting finding is that even though the model is usually clear about the authenticity of its output, it generates false content easily~\cite{Lying}. Given that, some methods try to shift model feature space during inference~\cite{ITI} to improve faithfulness. Nevertheless, designed for TruthfulQA~\cite{TruthfulQA}, these methods focus on LLM's internal knowledge and are unsuitable for long text generation like text summarization.

\section{Methodology}

We next describe our \textbf{C}ontrastive \textbf{P}reference \textbf{O}ptimization (\textbf{CPO}) with probing-based specific training method. The whole methodology can be divided into three parts: (1) Sentence-Level Data Collection, obtained by collecting summaries from the most common LLMs and designing an appropriate prompt to get accurate automatic annotation based on ChatGPT and GPT-4, (2) Contrastive Preference Optimization, where we encourage LLMs to generate with different propensities according to different instructions and adopt adversarial training to enhance LLMs' perception of their capabilities, and (3) Probing-based Specific Training, where we dynamically probe for the awareness of hallucinations and train the vulnerable modules to make up for the deficiency.

\subsection{Sentence-Level Data Collection}
\label{sec:DataCollection}
As mentioned in Section~\ref{sec:Introduction}, mistakes made by LLMs are much more subtle and challenging to detect or reproduce by previous methods. The previous studies use small models (smaller than 3B) to generate summaries whose distribution differs from those generated by LLMs. Hence, it is necessary to obtain a summarization dataset for LLMs. Considering that, we construct a dataset named LESSON containing summaries generated by current decoder-only LLMs, including GPT-family~\cite{opt}, GLM-family~\cite{glm, chatglm} and LLaMA-family~\cite{llama, llama2} models based on XSum~\cite{xsum} and CNN/DM~\cite{CNNDM}. More details about data collection are explained in Appendix~\ref{sec:appendixe}.

After obtaining the summaries from LLMs, we need to annotate their factual consistency. Most previous methods annotate the dataset at a sample-level~\cite{cliff}, which is quite improper for the LLMs' summaries because they are much longer, and only a few sentences in an inconsistent sample are false. On the other hand, it is challenging to get token-level annotation. \citet{Lying} prove that hallucinations in a sentence can be caused by qualifiers because an LLM generates a token at a time, and it "commits" to each token generated. Unfortunately, annotators usually neglect these qualifiers even if they eventually lead to factual mistakes. Given that, we choose sentence-level instead of token-level, which means any hallucination in a sentence will contribute to labeling the whole sentence as inconsistent.

\citet{DiverseRole} find LLMs highly consistent with human annotators, so we employ strong proprietary LLMs to collect sentence-level factual consistency annotation for these system-generated summaries. The core requirement for annotation and evaluation models revolves around how to define "factual consistency." In general summarization tasks, "factual consistency" refers to the alignment between summaries and the original articles (context). However, in News Summarization, where articles are drawn from actual news events, interpreting "factual consistency" as consistency with external, real-world facts can be misleading. Our focus is on maintaining consistency with the provided context, rather than external facts, as the original articles (which represent the extrinsic facts) are included in the annotation and evaluation prompts. Therefore, LLMs (annotators and evaluators) do not need to verify these external facts. Using ChatGPT and GPT-4, we experimented with various prompts, sentence numbering formats, and instructions to detect hallucinations and selected the most effective approach. The final prompts for summarization and annotation are listed in Appendix~\ref{sec:appendixa}. 

\begin{table}[h!]
    \centering
    \resizebox{0.9\linewidth}{!}{
    \begin{tabular}{cc} \hline  
         \makecell[c]{Methods}&  \makecell[c]{Balanced Accuracy}\\ \hline
         DAE&  63.75 \\
         QuestEval (mean)&  61.25 \\
         QuestEval (F1)&  53.75 \\
         SummaC-ZS (mean)&  53.33 \\
         SummaC-ZS (F1)&  59.58 \\
         SummaC-Conv (mean)&  51.25 \\
         SummaC-Conv (F1)&  56.67 \\
         QAFactEval (mean)&  53.33 \\
         QAFactEval (F1)&  64.16 \\
         \hline
         Ours&  \textbf{76.70} \\
         \hline 
    \end{tabular}
    }
    \caption{Results of traditional NLI and QA paradigm methods compared with ours on 160 samples under human evaluation.}
    \label{tab:compareeval}
\end{table}

\begin{table}[h!]
    \centering
    \resizebox{1.0\linewidth}{!}{
    \begin{tabular}{ccccc} \hline  
          \makecell[c]{Data\\Source}& \makecell[c]{Nums}&  \makecell[c]{Consistency\\(Pos/Neg)}&  \makecell[c]{Avg\\Words}& \makecell[c]{\#Avg\\Words}\\ \hline  
         XSum&  6166&  3521/2645&  34.96& 23.26 \\ \hline 
         CNN/DM&  4114&  2752/1362&  70.03& 51.84 \\ \hline 
    \end{tabular}
    }
    \caption{The statistics of the summaries of LESSON. More details can be found in Appendix~\ref{sec:appendixe}.}
    \label{tab:LESSON}
\end{table}

To check our annotation quality (the reliability of ChatGPT and GPT-4 as evaluators), we conducted human assessment on 160 LLMs' summaries to obtain the real labels and calculated Balanced Accuracy (BA) as~\citet{eval2020, summac, FEQA, Joshua, Asking} to measure the reliability:

\begin{equation}
\begin{aligned}
&TPR=TP/(TP+FN)\\
&TNR=TN/(TN+FP)\\
&BA=(TPR+TNR)/2
\label{BA}
\end{aligned}
\end{equation}

The experimental results are listed in Table~\ref{tab:compareeval}. Our auto-evaluation approach achieves 76.70\% accuracy while the best of previous ones only has a 64.16\% accuracy, proving ours has a much higher consistency with humanity on LLMs' summaries. After annotating all the summaries generated by LLMs previously, we get the final dataset - \textbf{LESSON}. The statistics are shown in Table~\ref{tab:LESSON} and more details can be found in Appendix~\ref{sec:appendixe}. The average length of summaries generated by LLMs is much longer than that of reference summaries, proving the big gap between the current view based on strong LLMs and the previous one.

\subsection{Contrastive Preference Optimization}
\label{sec:decent}

\begin{figure*}[t!]
	\centering
	\includegraphics[width=1.0\linewidth]{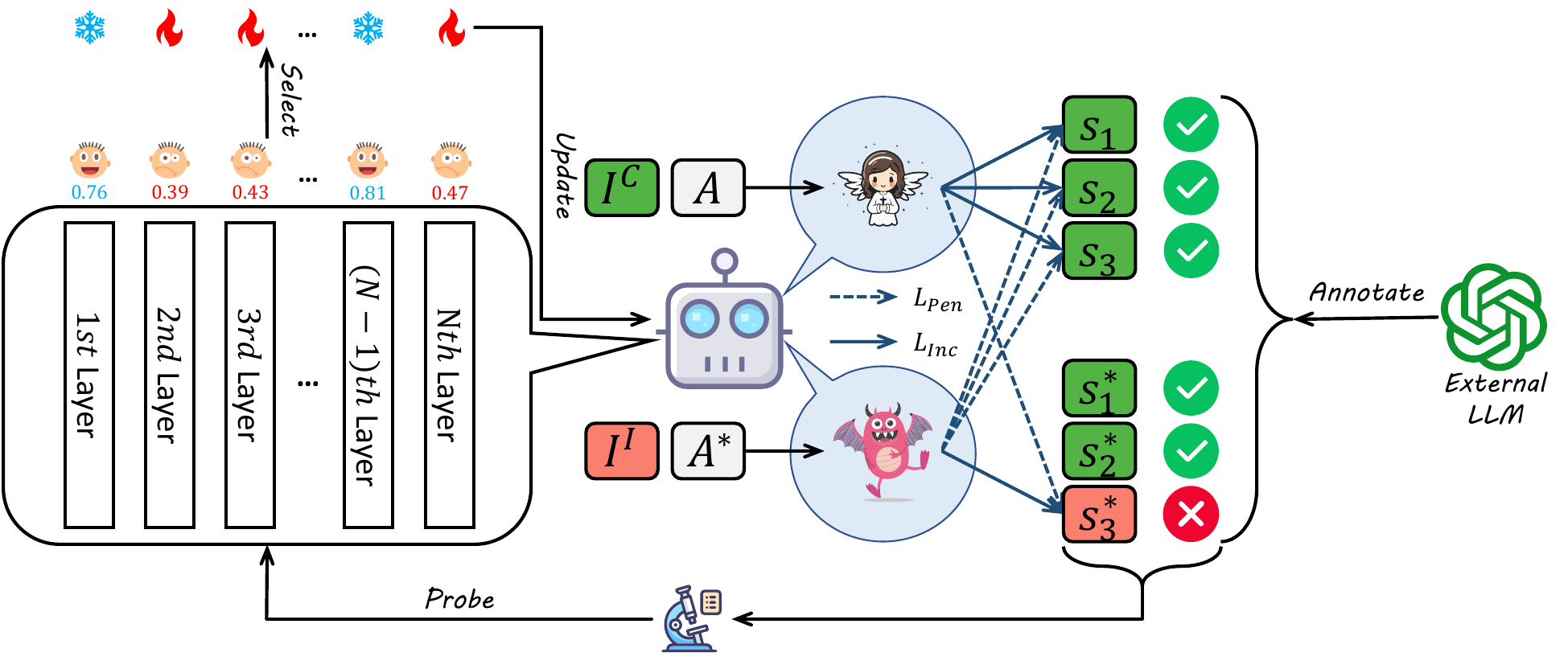}
    \caption{The diagram of our method. Based on LESSON annotated by ChatGPT and GPT-4, we adopt Incentive Loss and Penalty Loss to optimize LLMs' contrastive preferences. Meanwhile, we dynamically calculate the probing scores of each layer and employ probing specific training to select weak layers to remedy their insensitivity.}
    \label{Fig:mainmodel}
\end{figure*}

Having the dataset with sentence-level annotation, we can optimize contrastive preferences in finer-grain. \citet{DisentQA} find that LLMs have parameterized and contextual knowledge, which results in different generation propensities. In summarization, we expect the LLMs to focus on the original articles but not their own parameterized knowledge. Hence, we must decouple the LLMs' preferences to be faithful to the context and imagine with internal knowledge. As shown in Figure~\ref{Fig:mainmodel}, we design two instructions for the two preferences. The contextual instruction named $I^C$ and the internal one named $I^I$ are listed in Appendix~\ref{sec:appendixa}.

Before training, the original model does not know how to meet the demands of the two contrastive instructions and just write summaries with their strong generation capabilities. Hence, we design an Incentive Loss to encourage the model to follow the instructions. Given a summary $S$ consists of $n$ words $S=[w_1,w_2,...,w_n]$ annotated with the label set $L=[l_1,l_2,...,l_n]$, we can divide $S$ into $S^{+}=\{w_i|l_i=1,i\in [1,n]\}$ and $S^{-}=\{w_j|l_j=0,j\in [1,n]\}$. Given that, Incentive Loss is defined for consistent and inconsistent summaries, respectively:
\begin{equation}
\begin{aligned}
L_{Incentive}&= Y\sum\limits_{w_i \in S^{+}} \log P(w_i|w_{<i};I^C;\Theta)\\
+(1-&Y)\sum\limits_{w_j \in S^{-}} \log P(w_j|w_{<j};I^I;\Theta)
\end{aligned}
\label{Incentive}
\end{equation}
where Y denotes the faithfulness of the summary $S$. $Y=1$ only if all the sentences in $S$ are completely true and $Y=0$ as long as any sentence is inconsistent:
\begin{equation}
\begin{aligned}
&Y=
\begin{cases}
1 \quad if \quad S^{-}=\varnothing \\
0 \quad otherwise
\end{cases}
\end{aligned}
\label{Y}
\end{equation}

Given $I^I$, although there are hallucinations in the generated summary, we still encourage the behavior because the model executes the instruction precisely. It is worth noting that only the hallucinatory sentences in the inconsistent summary are taken into consideration while calculating $L_{Incentive}$, because those factually consistent sentences mixed with them are not supposed to be proper output of $I^I$.

Apart from teaching the model what it should do, we also teach it what it should not do. In other words, we need to penalize disobeying an instruction. We do not expect the model to generate inconsistent sentences with $I^C$ or consistent sentences with $I^I$. Hence, the Penalty Loss for adversarial training is defined as:
\begin{equation}
\begin{aligned}
&L_{Penalty}= Y\sum\limits_{w_i \in S^{+}} \log (1-P(w_i|w_{<i};I^I;\Theta))\\
&+(1-Y)\sum\limits_{w_j \in S^{-}} \log (1-P(w_j|w_{<j};I^C;\Theta))
\end{aligned}
\label{Penalty}
\end{equation}

Similarly, under $I^C$, we only punish the generation of false sentences. As for the right sentences in factually incorrect summaries, we neither incent nor penalize them because these sentences indeed follow $I^C$.

Finally, the total training loss can be written as:
\begin{equation}
\begin{aligned}
&L= L_{Incentive} + \alpha L_{Penalty}
\end{aligned}
\label{TOTAL}
\end{equation}
where $\alpha$ is the hyperparameter to balance the strength of punishment and the training objective.

\subsection{Probing-based Specific Training}
~\citet{mario} find that most of the trainable parameters can be directly discarded without significantly affecting the capabilities of SFT LLMs. In other words, full parameter training for LLMs is usually unnecessary, and finding more "profitable" modules is crucial for conducting more specific and efficient training. Section~\ref{sec:decent} has explained how CPO teaches LLMs to follow different instructions to generate faithful and fake summaries. Still, the performance of instruction following relies on distinguishing between consistent and inconsistent content. Given that, we conduct probing-based training to specifically train those modules being unclear about the differences between contrastive summaries. Specifically, we utilize a probing set named DeFacto~\cite{feedback}, where each article has a correct summary $S^C$ and an incorrect one $S^I$. For each summary $S$ ($S^C$ or $S^I$) of a length $T$, we construct a probing prompt by concatenating it with the corresponding article in the format of $I^C$ and feed it into the model $\boldsymbol{M}$, whose output shape is $(N, L, D)$ (the number of layers, length of sequence, and hidden size). Given that, we can obtain the hidden states of $\boldsymbol{M}$'s different layers:

\begin{equation}
\begin{aligned}
&F = [f^1,f^2,...,f^N] = \boldsymbol{M}(I^C;S)\\
&f^u = [f^u_1,f^u_2,...,f^u_T] \quad u \in [1,N]
\end{aligned}
\label{getfeature}
\end{equation}
 
Then, we use the hidden state of the last token at layer $u$ to train a binary linear classifier $\phi$ which identifies whether the summary is consistent with the source text. The training loss is:
\begin{equation}
\begin{aligned}
&L_{Probing} = -\sum_{S} Y_{S} \log \phi(f^u_T)\\
&+(1-Y_{S}) \log (1-\phi(f^u_T))
\end{aligned}
\label{trainprobe}
\end{equation}

Finally, the classifiers' accuracy reflects the layer $u$'s capability to distinguish the factuality. As shown in Figure~\ref{Fig:compare}, the intermediate layers usually have a clear sense of whether the summary is accurate, but the bottom and top layers do not have the same ability, which suggests the weakness of these layers in understanding and following instructions precisely. Intuitively, the distinguishing capacity is closely related to the final effect of CPO. We expect the model to be faithful to the context given the contextual instruction $I^C$ on the premise of following instructions precisely, which requires awareness of its generation's factuality. Considering that, we dynamically probe and select the top-$k$ worst layers to train so that the training stage focuses on the model's weakness without interference from other layers. The overall training process is listed in Algorithm~\ref{train-process} and various selections of layers are discussed in Appendix~\ref{sec:appendixf}.

\begin{algorithm}[h!]
\caption{\textbf{Training Process}}
\KwIn{training set $D_{t}$, probing set $D_{p}$, language model $\boldsymbol{M}$, training epochs $E$.}
\For{$i = 1,,2...,E$}{
    Probe the model $\boldsymbol{M}$ on $D_{p}$ to obtain probing scores $A$.\\
    Select the $k$ worst layers according to the accuracy.\\
    Train the $k$ layers of $\boldsymbol{M}$ with $L$ on $D_{t}$.
    }
\Return $\boldsymbol{M}$
\label{train-process}
\end{algorithm}

\section{Experiments}
We conduct extensive experiments to verify the effectiveness of our proposed model DECENT and analyze it using ablation studies, case studies, and visualization results. In this section, we attempt to answer the following research questions: \textbf{RQ1:} Does CPO improve LLM's summarization factual consistency? \textbf{RQ2:} Does CPO decouple LLMs' different propensities successfully? \textbf{RQ3:} Does Probing-based Specific Training fill the gaps? \textbf{RQ4:} Is CPO better than other training strategies?

\subsection{Experimental Details}

\paragraph{Datasets}
To evaluate the effectiveness of our model, we conduct training experiments on LESSON (train-validation split is 9:1), the construction and statistics of which have already been explained above. Each sample is generated by a certain LLM and has a sentence-level annotation. The auto evaluation is based on ChatGPT and GPT-4. The prompt is the same as the one we use to collect factual consistency annotation, whose reliability has been proven in Section~\ref{sec:DataCollection}. As for the human evaluation, we collect 300 articles from the test set of original datasets, and the model-generated summaries are assigned to human annotators after shuffling. Each summary is evaluated by two workers while masking its source (the workers do not know whether the summary comes from CPO+PST or original backbones). The evaluation criterion is discussed in Appendix~\ref{sec:appendixg}.

\paragraph{Backbones} To initialize the summarization model, we use ChatGLM2-6B, LLaMA2-7B-chat, Koala-7B, Tulu-7B, Vicuna-7B, and BLOOMZ-7B. Noteworthily, we focus on how to improve LLM's factual consistency for summarization and do not expect to instruction-tune them from the beginning, so we choose these models as backbones because of their ability to understand and execute the instructions for summarization, despite lots of hallucinations in their generation. To prove the effectiveness of our approach more comprehensively, we also conducted the experiment on OPT-6.7B and Pythia-12B, which are only pre-trained without any extra instruction tuning.

\paragraph{Experimental Settings} We conducted parallel training on 8*NVIDIA A100 80G for all backbones. The batch size is set to 8, and the number of epochs is set to 5. The learning rate is 1e-5, and the weight decay is 3e-7. WarmupLR scheduler is also used with a warmup ratio of 0.2. As for hyperparameters, we set $\alpha$ as 0.05.

\subsection{\textbf{RQ1:} Does CPO with PST improve LLM's factual consistency?}
\label{sec:faccon}
As shown in Table~\ref{tab:mainres} and Figure~\ref{Fig:win}, \textbf{CPO+PST significantly improves the LLMs' factual consistency under both automatic and human evaluation}. ChatGPT and GPT-4 may generate different opinions on the same article with their own standards and preferences, but our method performs well under both assessment systems. CPO+PST performs better than CPO (full-parameter fine-tuning), proving CPO benefits from specific training. The human evaluation for other qualities of summaries can be found in Appendix~\ref{sec:appendixh}, indicating they are not sacrificed to improve faithfulness.

It's worth noting that nearly all these models are pre-trained or instruction-tuned on CNN/DM, so their original performance on CNN/DM (in-domain) is much better than that on XSum (out-of-domain). For example, ChatGLM2 and LLaMA2 are tuned on CNN/DM and OpenAI Summarize~\cite{openaisum} (a variant of CNN/DM with human feedback), respectively. \textbf{The extra SFT can easily lead to overfitting because they have been thoroughly trained in the domain}, and PST alleviates it to some extent.

OPT-6.7B and Pythia-12B have yet to be instruction-tuned, resulting in their inability to understand the instructions(they often output invalid content like URLs and continuations of the original article). However, after training by CPO+PST, they can achieve a competitive performance compared to the others having been instruction-tuned on large-scaled in-domain corpora, which indicates \textbf{CPO teaches the models to summarize precisely with a pretty small amount of data, and models' essential instruction understanding capacities do not constrain its effectiveness}.

\begin{table}[h!]
    \centering
    \resizebox{1.0\linewidth}{!}{
    \begin{tabular}{lccccc} \hline  
          \multicolumn{2}{c}{\multirow{2}{*}{Models}}&  \multicolumn{2}{c}{XSum}& \multicolumn{2}{c}{CNN/DM}\\
          \multicolumn{2}{c}{}&  ChatGPT&  GPT-4&  ChatGPT&  GPT-4\\ \hline 
          \multirow{3}{*}{ChatGLM2 (6B)}&vanilla&  0.47&  0.43&  \textbf{0.88}&  0.87 \\
          &CPO&  0.52&  \textbf{0.45}&  0.81&  0.83 \\
          &CPO+PST&  \textbf{0.55}&  0.42&  0.83&  \textbf{0.88} \\
          \hline
          \multirow{3}{*}{LLaMA2 (7B)}&vanilla&  0.76&  0.78&  \textbf{0.89}&  \textbf{0.88} \\
          &CPO&  0.81&  0.79&  0.81&  0.82 \\
          &CPO+PST&  \textbf{0.84}&  \textbf{0.89}&  0.85&  0.84 \\
         \hline 
         \multirow{3}{*}{Koala (7B)}&vanilla&  0.58&  0.67&  0.64&  0.82 \\
          &CPO&  0.75&  0.72&  0.76&  \textbf{0.85} \\
          &CPO+PST&  \textbf{0.83}&  \textbf{0.81}&  \textbf{0.78}&  0.84 \\
         \hline 
         \multirow{3}{*}{Tulu (7B)}&vanilla&  0.72&  0.71&  0.78&  0.80 \\
          &CPO&  0.76&  0.72&  0.80&  0.82 \\
          &CPO+PST&  \textbf{0.82}&  \textbf{0.81}&  \textbf{0.91}&  \textbf{0.86} \\
         \hline 
          \multirow{3}{*}{Vicuna (7B)}&vanilla&  0.61&  0.71&  0.58&  0.56 \\
          &CPO&  \textbf{0.84}&  0.78&  \textbf{0.77}&  0.72 \\
          &CPO+PST&  0.81&  \textbf{0.84}&  0.74&  \textbf{0.76} \\
         \hline 
         \multirow{3}{*}{BLOOMZ (7B)}&vanilla&  0.74&  0.54&  0.74&  0.80 \\
          &CPO&  0.76&  0.59&  0.76&  \textbf{0.84} \\
          &CPO+PST&  \textbf{0.78}&  \textbf{0.76}&  \textbf{0.85}&  0.83 \\
          \hline
         \multirow{3}{*}{OPT (6.7B)}&vanilla&  0.12&  0.22&  0.62&  0.53 \\
          &CPO&  0.71&  0.71&  \textbf{0.88}&  0.84 \\
          &CPO+PST&  \textbf{0.80}&  \textbf{0.86}&  0.84&  \textbf{0.92} \\
         \hline 
         \multirow{3}{*}{Pythia (12B)}&vanilla&  0.50&  0.35&  0.53&  0.34 \\
         &CPO&  0.71&  0.67&  0.74&  0.67 \\
          &CPO+PST&  \textbf{0.85}&  \textbf{0.72}&  \textbf{0.87}&  \textbf{0.86} \\
         \hline 
    \end{tabular}
    }
    \caption{Overall factual consistency.}
    \label{tab:mainres}
\end{table}

\begin{figure}[h!]
	\centering
	\includegraphics[width=0.8\linewidth]{pic/win.pdf}
    \caption{The win rate of CPO+PST on factual consistency under human evaluation.}
    \label{Fig:win}
\end{figure}

\subsection{\textbf{RQ2:} Does CPO decouple LLMs' different propensities successfully?}
\label{sec:doesdecouple}
In this section, we evaluate the effectiveness of CPO from the perspective of following both $I^C$ and $I^I$. As Table~\ref{tab:onecase} shows, given the same article, there's a visible difference between their corresponding generations. \textbf{$I^C$ generates the factually consistent summary, while $I^I$ summarizes with hallucinations}. Interestingly, these hallucinations are not completely irrelevant to the source article. On the contrary, they are generated through the "reasonable" imagination and inference of the model, which makes the wrong summary seem like an adaptation of the original article. For example, $I^I$ can write what will happen as what has already happened, add some fake details, overgeneralize, and disguise the replacement of concepts. More examples can be found in Appendix~\ref{sec:appendixb}.

\begin{table}[t!]
    \centering
    \resizebox{1.0\linewidth}{!}{
    \begin{tabular}{p{1.0\columnwidth}}\hline
         \textcolor{blue}{Source}: The stone memorial is on the banks of Llyn Egnant - one of the famous Teifi pools - near the village of Ffair Rhos in Ceredigion. He died last year at the age of 86. A prominent figure in the Welsh fishing community, Mr Morgan once took the former US President Jimmy Carter on a fishing trip in mid Wales and they became great friends.  The stone memorial was unveiled by his widow, Julia Morgan. He was described as the "grand-daddy of game angling in Wales" by Cheryl Bulman, of Tregaron Angling Association, which is celebrating its centenary year.  She said that Mr Morgan, the Teifi River and Tregaron Angling Association were "intrinsically linked".\\
         \textcolor{blue}{$I^C$}: A stone memorial has been unveiled on the banks of a Welsh lake to commemorate the life of a prominent fisherman who died last year.\\
         \textcolor{blue}{$I^I$}: A stone memorial for \textcolor{red}{a former fishing guide} has been unveiled in Wales. The memorial, located on the banks of a river in the Teifi Valley, \textcolor{red}{was erected by his widow} and dedicated to his memory.\\
         \hline
    \end{tabular}
    }
    \caption{Different outputs of Vicuna (7B) under $I_{com}$ and $I_{emb}$ after being trained by CPO. The words in \textcolor{red}{red} are hallucinations.}
    \label{tab:onecase}
\end{table}

\begin{figure*}[t!]
	\centering
	\includegraphics[width=1.0\linewidth]{pic/compare.pdf}
    \caption{The head-level probing results. Darker green means higher accuracy.}
    \label{Fig:compare}
\end{figure*}

\subsection{\textbf{RQ3:} Does Probing-based Specific Training fill the gaps?}
\label{sec:PET}

\begin{figure}[t!]
	\centering
	\includegraphics[width=1.0\linewidth]{pic/training.pdf}
    \caption{The factual consistency of the checkpoints under different training epochs on XSum.}
    \label{Fig:training}
\end{figure}

For a fine-grained observation, we conduct probing tests for each attention head of the models. Figure~\ref{Fig:compare} indicates that \textbf{CPO remarkably enhances LLM's discernment capacity}, especially for the bottom and top layers, which are originally insensitive. The statistics of probing scores are listed in Appendix~\ref{sec:appendixc}, which are aligned with the visualization, proving CPO enables LLMs to distinguish between consistent and inconsistent summaries more clearly.

In addition, we also find \textbf{probing-based specific training is much more stable than full-parameter fine-tuning}. The variations in factual consistency of different checkpoints are shown in Figure~\ref{Fig:training}. The performance of training without PST first peaks and then declines rapidly, while training with PST maintains a high consistency, which indicates that full-parameter fine-tuning is easily affected by unnecessary training. Still, PST potentially makes the training stage more targeted.

\begin{table}[t!]
    \centering
    \resizebox{1.0\linewidth}{!}{
    \begin{tabular}{ccccccc} \hline
         \multirow{2}{*}{Models}&  \multicolumn{2}{c}{Vicuna (7B)}&  \multicolumn{2}{c}{LLaMA2 (7B)}&  \multicolumn{2}{c}{BLOOMZ (7B)} \\ 
         {}&  \makecell[c]{ChatGPT}&  \makecell[c]{GPT-4}& \makecell[c]{ChatGPT}&  \makecell[c]{GPT-4}& \makecell[c]{ChatGPT}&  \makecell[c]{GPT-4} \\ \hline
         Vanilla&  0.61&  0.71&  0.76& 0.78& 0.74& 0.80 \\
         SFT&  0.75&  0.76& 0.77& 0.74& 0.77& 0.78 \\
         SFT+LoRA&  0.63& 0.72& 0.77& 0.76& 0.74& 0.78 \\
         Contrastive Learning&  0.81&  0.66& 0.76& 0.75& 0.75& 0.73 \\
         Unlikelihood Optimizing&  0.71&  0.74& 0.77& 0.65& 0.78& 0.72 \\
         Decoupling&  0.77&  0.72& 0.79& 0.67& 0.74& 0.71 \\
         PPO&  0.75&  0.81& 0.70& 0.78& 0.74& 0.72 \\
         DPO&  0.80&  0.74& 0.83& \textbf{0.89}& 0.82& 0.73 \\
         DPO+LoRA&  0.72& 0.75& 0.78& 0.80& 0.77& 0.76 \\
         CPO&  \textbf{0.84}&  0.78& 0.81& 0.79& 0.76& \textbf{0.84} \\
         CPO+LoRA&  0.78& 0.74& 0.77& 0.82& 0.80& 0.78 \\
         CPO+PST&  0.81&  \textbf{0.84}& \textbf{0.84}& \textbf{0.89}& \textbf{0.85}& 0.83 \\
         \hline 
    \end{tabular}
    }
    \caption{CPO+PST compared with current strong training strategies for LLMs' summarization.}
    \label{tab:abalation}
\end{table}

\subsection{\textbf{RQ4:} Is CPO better than other training strategies?}
We try different training strategies and compare them with CPO in Table~\ref{tab:abalation}, including:

\textbf{SFT}: Only train the model on high-quality positive samples including instructions and responses (Incent the output of truthful summaries).
\textbf{Contrastive Learning}: Use the Mixed-Contrast Loss~\cite{aaaicontrast} for negative samples in SFT (Incent the output of truthful summaries and penalize the hallucinations).
\textbf{Unlikelihood Optimizing}: Introduce Unlikelihood Loss~\cite{unlikelihood} to training (Incent the output of truthful summaries and penalize the hallucinations).
\textbf{Decoupling}: Decouple the models' abilities (Incent both truthful and false summaries as long as they are consistent with the instructions).
\textbf{PPO}: Apply reinforcement learning with a reward model~\cite{PPO} (Proximal policy optimization).
\textbf{DPO}: Replace the reward scores in PPO with chosen-rejected pairs~\cite{DPO} (Direct preference optimization). The core idea of DPO is to directly optimize the language model (LM) to fit human preferences, rather than first fitting a reward model and then optimizing it using reinforcement learning.
\textbf{CPO}: Adversarially decouple the model's propensities to generate faithful and fake content (Incent both truthful and false summaries and penalize disobeying the instructions).
\textbf{CPO+PST}: CPO with probing-based efficient training.

All these training strategies improve the performance of the vanilla model to different degrees. In general, the effect of the incentive-only paradigm is more stable than that of the incentive-with-penalty paradigm, which indicates that \textbf{the punishment for generating hallucinations may affect the stability of the training process}. Outperforming SFT and Decoupling, CPO and CPO+PST get the best factual consistency on most tests, indicating that adversarially training is significant and PST makes it possible to train just several layers of a model to achieve a competitive or even better performance compared with full-parameter fine-tuning.

Additionally, we compared these methods with RL-based approaches. While PPO depends on reward models and DPO requires paired data, CPO offers a more flexible and accessible training paradigm through SFT, as opposed to the traditional RL framework. The higher consistency achieved by CPO+PST, in comparison to PPO and DPO, suggests that \textbf{assessing which summary is superior at the document level may be too complex for training}. For a more comprehensive comparison, we evaluated CPO+PST against LoRA tuning~\cite{LORA}, which also refrains from updating the full set of parameters. However, the restricted number of trainable parameters in LoRA limits the method's overall effectiveness.

\section{Conclusion}
This paper points out the problems of applying previous methods for summarization factual consistency to LLMs. We construct a summarization dataset (LESSON) and propose Contrastive
Preference Optimization with Probing-based Specific Training to improve factual consistency. The experimental results demonstrate the effectiveness of our method on the most common LLMs. We expect our work will direct more scholarly attention to constructing new datasets and enhancing factual consistency from the perspective of LLMs.

\section*{Limitations}
In this paper, we propose Contrastive
Preference Optimization with Probing-based Specific Training. Although CPO+PST significantly improves the factual consistency of all backbones, unnecessary training can easily affect its performance, especially on the in-domain dataset. As discussed in Section~\ref{sec:faccon} and Appendix~\ref{sec:appendixf}, selecting an appropriate value for hyperparameters is essential.

\section*{Acknowledgements}
We thank the anonymous reviewers for their valuable feedback. This work was partially funded by the National Natural Science Foundation of China (Grant No. 62272173), the Natural Science Foundation of Guangdong Province (Grant Nos. 2024A1515010089, 2022A1515010179), the Science and Technology Planning Project of Guangdong Province (Grant No. 2023A0505050106), the Fundamental Research Funds for the Central Universities (Grant No. 2024ZYGXZR104), and supported by Alibaba Group through the Alibaba Research Intern Program.

\appendix

\section{The details about data collection.}
\label{sec:appendixe}

\begin{table}[t!]
    \centering
    \resizebox{1.0\linewidth}{!}{
    \begin{tabular}{p{1.0\columnwidth}}\hline
         \textcolor{blue}{Prompt}: \textcolor{darkgray}{Answer which sentences in the summary are not consistent with the corresponding article. Provide the answer in JSON format like this: \{"inconsistent\_sentence": [indexes of inconsistent sentences], "consistent\_sentence": [indexes of consistent sentence]\}}\\
         \textcolor{darkgray}{<article>}\\
         \textcolor{violet}{Like last year big-spending Mazembe drop into the Confederation Cup after exiting the Champions League before the group stage. The Congolese, who have are five-time African champions, will be hoping to appoint a new coach before the two matches in April to decide who advances group stage. This after the club announced that Frenchman Thierry Froger had left by mutual consent after just over one month in charge. Mazembe said he had not achieved his goal of reaching the Champions League quarter-finals after they Mazembe lost to Zimbabwe's CAPS United on the away goals rule in the round of 32. Two-time African champions Kabylie beat Congo's Etoile to reach the play-offs. Tuesday's draw for pits losers from Champions League against second-round winners from the Confederation Cup to decide who reaches the expanded group stage. This year's tournament will feature 16 teams in four pools up from eight sides in previous years.}\\
         \textcolor{darkgray}{</article>}\\
         \textcolor{darkgray}{<summary>}\\
         \textcolor{violet}{(0)    The Confederation Cup draw} \textcolor{red}{has taken place,} \textcolor{violet}{with 16 teams split into four groups.}\\
         \textcolor{violet}{(1)}    \textcolor{red}{The Congolense will face off against the second-round winners of the Confederation Cup.}\\
         \textcolor{darkgray}{</summary>}\\
         \textcolor{blue}{ChatGPT's response}: \{"inconsistent\_sentence": [0, 1],"consistent\_sentence": []\}\\
         \textcolor{blue}{GPT-4's response}: \{"inconsistent\_sentence": [1],"consistent\_sentence": [0]\}\\
         \hline
    \end{tabular}
    }
    \caption{An example of how to use the annotation prompt. The words in \textcolor{red}{red} are hallucinations.}
    \label{tab:promptexp}
\end{table}

\begin{table}[t!]
    \centering
    \resizebox{0.8\linewidth}{!}{
    \begin{tabular}{ccccc} \hline
         \multirow{2}{*}{Models}&  \multicolumn{2}{c}{XSum}&  \multicolumn{2}{c}{CNN/DM} \\ 
         {}&  \makecell[c]{POS}&  \makecell[c]{NEG}& \makecell[c]{POS}&  \makecell[c]{NEG} \\ \hline
         chatglm-6b& 84& 374& -& - \\
         koala-7b& 187& 216& 108& 143 \\
         koala-13b& 210& 207& 77& 142 \\
         vicuna-7b& 136& 164& 41& 69 \\
         vicuna-13b& 78& 68& 94& 87 \\
         llama-7b& 297& 183& 167& 237 \\
         llama-13b& 283& 197& 170& 244 \\
         tulu-7b& 222& 228& 213& 122 \\
         tulu-13b& 272& 206& 209& 151 \\
         bloom-7b& 175& 308& -& - \\
         chatgpt/gpt-4& 1577& 494& 1673& 167 \\
         \hline 
    \end{tabular}
    }
    \caption{The composition of LESSON. POS and NEG indicate the number of positive summaries (all the sentences are factually consistent) and the number of negative summaries (at least one sentence is inconsistent with the original text).}
    \label{tab:LESSONdetails}
\end{table}

In this section, we talk about the details of data collection. The models used to generate summaries come from GPT-family~\cite{opt}, GLM-family~\cite{glm, chatglm} and LLaMA-family~\cite{llama, llama2, longpre2023flan, köpf2023openassistant, peng2023instruction, codealpaca}, including BLOOMZ-7B, ChatGPT, GPT-4, ChatGLM-6B, LLaMA2-7B-chat, LLaMA2-13B-chat, Koala-7B, Koala-13B, Tulu-7B, Tulu-13B, Vicuna-7B, and Vicuna-13B.

Even though we explicitly inform the models to "write a summary consistent with the above article", they still make many factual mistakes. Noteworthily, some summaries exceed the max length limitation, which may cause errors during the training, so we delete them from the dataset. Otherwise, some summaries even involve errors other than hallucinations, including sentence fragments and mixtures of multiple languages, which is harmful to the SFT stage. Given that, we remove these poor-quality ones according to the heuristic rules including:
\begin{itemize}

\item Remove the summaries containing sentence fragments. Vicuna and Tulu sometimes generate the fragmentary summaries. A complete sentence should end with a punctuation mark so we can remove those summaries failing to meet the condition.

\item Remove the summaries containing mixtures of multiple languages. ChatGLM2 and LlaMA2 are pre-trained on multilingual data so they sometimes output mixtures of multiple languages. We use lid.176.bin from FastText Language Identification to remove those non-English summaries.

\item Remove the summaries containing strange symbols. GPT-4 and ChatGPT can output some strange symbols like emojis, format symbols and so on. We remove these data by regex matches.
\end{itemize}

It is worthy noted that the ChatGPT's response may be different from GPT-4's, so we fetch the union of their annotations to get a high recall. In other words, the method will be pretty strict with the summary and sometimes may regard some true sentences as false ones according to the annotators' preferences. Still, it is acceptable for the training stage because that forces the model to learn a more rigorous expression.

After annotating with ChatGPT and GPT-4, we get the final dataset - LESSON. The statistics of LESSON are shown in Table~\ref{tab:LESSON} and Table~\ref{tab:LESSONdetails}.

\section{The details of prompts.}
\label{sec:appendixa}

This section introduces the prompts to collect annotations and conduct adversarial decoupling.

The prompt to \textbf{collect sentence-level annotations} is:\\
\textcolor{darkgray}{Answer which sentences in the summary are not consistent with the corresponding article. Provide the answer in JSON format like this: \{"inconsistent\_sentence": [indexes of inconsistent sentences], "consistent\_sentence": [indexes of consistent sentence]\} \\<article> \textcolor{violet}{[ARTICLE]} </article> \\<summary> \textcolor{violet}{[SUMMARY]} </summary>}

As shown in Table~\ref{tab:promptexp}, we split the summary into sentences and add indexes in front of them. Otherwise, we find numbering the sentences from zero is much better than numbering from one. In this way, we can get annotations from ChatGPT and GPT-4 in JSON format. However, the ChatGPT's response may be different from GPT-4's. Each of ChatGPT and GPT-4 only has a balanced accuracy less than 70\%, but the union of their annotation can reach 76.7\%, which means they have minor disagreements. Considering that, we fetch the union of their annotations to get a high recall. In other words, the method will be pretty strict with the summary and try to detect each hallucination. Sometimes, it will regard some true sentences as false ones according to their own preferences. Still, it is acceptable for the training stage because that forces the model to learn a more rigorous expression. In the final evaluation, we still use ChatGPT and GPT-4 separately because they are two different sets of assessment systems anyway.

As for the seed prompts in Section~\ref{sec:decent}, the \textbf{internal} instruction seed named as $I^I$ is:\\
\textcolor{darkgray}{Article: \textcolor{violet}{[ARTICLE]}. Write a summary \textcolor{red}{inconsistent} with the above article in no more than 40 words:}\\
and the \textbf{contextual} instruction seed named as $I^C$ is:\\
\textcolor{darkgray}{Article: \textcolor{violet}{[ARTICLE]}. Write a summary \textcolor{red}{consistent} with the above article in no more than 40 words:}\\
Noteworthily, the instruction $I_{com}$ is also used in Section~\ref{sec:DataCollection} to collect real model-generated summaries. Certainly, the LLMs are not aware of how to meet "\textbf{consistent}" and "\textbf{inconsistent}" before training, so there are still lots of hallucinations in the original summaries.

\section{The number of trainable layers.}
\label{sec:appendixf}

\begin{table}[h!]
    \centering
    \resizebox{0.8\linewidth}{!}{
    \begin{tabular}{ccc} \hline  
         \makecell[c]{Models}&  \makecell[c]{ChatGPT}&  \makecell[c]{GPT-4} \\ \hline
         CPO+PST (k=2)&  0.84&  0.79 \\
         CPO+PST (k=4)&  0.81&  0.84 \\
         CPO+PST (k=8)&  0.76&  0.82 \\
         CPO+PST (k=16)&  0.74&  0.77 \\
         CPO+Random (k=4)&  0.79&  0.80 \\
         CPO+Best (k=4)&  0.81&  0.72 \\
         \hline 
    \end{tabular}
    }
    \caption{Results of Vicuna (7B) trained with different training strategies.}
    \label{tab:select}
\end{table}

\begin{table*}[t!]
    \centering
    \resizebox{1.0\linewidth}{!}{
    \begin{tabular}{p{2.0\columnwidth}}\hline
         \textcolor{blue}{Source}: Its futuristic curves fit nicely with Tokyo's Olympic slogan: "Discover Tomorrow." But it comes at a huge price: more than \$1.5 billion.  That is just a small part of the \$8 billion that will be spent on refurbishing old stadiums and building new ones. So it's all good news for Japan's construction companies, who have long been suckled on the teat of government spending. But all this money going it to Tokyo may be less good news for other parts of the country, particularly...\\
         \textcolor{blue}{$I^C$}: The announcement of Tokyo as the winning city for the Olympic Games has caused excitement among Japanese people, but also raises questions about the impact on the country's economy...\\
         \textcolor{blue}{$I^I$}: The Japanese government \textcolor{red}{has spent more than \$8 billion} on refurbishing and building new stadiums for the Olympic Games, which will be held in Tokyo...\\
         \hline
         \textcolor{blue}{Source}: Lam, 28, joined the club in 2014 and, despite missing out on promotion to the Premiership, the Samoa forward has ignored interest elsewhere to re-sign. He says the challenge of getting the club into the top flight was one he could not walk away from after their play-off loss to Worcester in May. "I feel I've got unfinished business here," Lam told BBC Radio Bristol... I got a couple of approaches from other clubs but I couldn't see myself leaving having not helped them get promoted," he said...\\
         \textcolor{blue}{$I^C$}: Lam re-signs with Bristol after play-off loss, says he couldn't leave the club without helping them get promoted.\\
         \textcolor{blue}{$I^I$}: Lam, a forward for Bristol Rugby, has signed a new contract with the club, \textcolor{red}{committing to the team until 2022}.\\
         \hline
         \textcolor{blue}{Source}: Shocked by the incident and other similar attacks on Africans in India, independent photographer Mahesh Shantaram began documenting the lives of Africans living in India. Beginning with Bangalore, Shantaram travelled to the cities of Jaipur, Delhi and Manipal, choosing to focus on students, as they are a small and vulnerable group. Shantaram's series of intimate portraits is part of an upcoming exhibition organised by Tasveer...\\
         \textcolor{blue}{$I^C$}: A photographer has captured images of Africans in India, highlighting the challenges faced by Africans in the country.\\
         \textcolor{blue}{$I^I$}: An exhibition of photographs by Mahesh Shantar \textcolor{red}{is being held in Bangalore} to showcase the lives of \textcolor{red}{African students in India}.\\
         \hline
    \end{tabular}
    }
    \caption{Different outputs of Vicuna (7B) under different instructions after training. The words in \textcolor{red}{red} are hallucinations.}
    \label{tab:cases}
\end{table*}

In this section, we show the influence of the selection of trainable layers on the final performance. As Table~\ref{tab:select} shows, different settings for $k$ vary in the final effectiveness. A smaller $k$ makes the model inadequately trained, while a larger one causes overfitting. So, it is crucial to flexibly choose different values of k according to different backbones, especially for those having trained on in-domain datasets. On the other hand, randomly selecting the trainable layers or selecting the ones with the highest probing scores do not behave better than PST, which indicates the significance of training the weak layers.

\begin{figure}[h!]
	\centering
	\includegraphics[width=0.8\linewidth]{pic/winc.pdf}
    \caption{The win rate of CPO+PST on coherence under human evaluation.}
    \label{Fig:winc}
\end{figure}

\begin{figure}[h!]
	\centering
	\includegraphics[width=0.8\linewidth]{pic/wincover.pdf}
    \caption{The win rate of CPO+PST on content coverage under human evaluation.}
    \label{Fig:wincover}
\end{figure}

\section{The human evaluation criterion.}
\label{sec:appendixg}

In this section, we show how to conduct human evaluation. The human annotators are asked to evaluate the summaries from the perspective of factual consistency. Each article has two corresponding summaries (the original backbone generates one, and CPO+PST generates the other), and the workers must annotate the index of the better one. The source of summaries is masked to make a fair competition, which means the workers will not know where the summary comes from.

Table~\ref{tab:evalcases} shows three cases. In the first case, Summary 1 contains hallucinations, but Summary 2 does not, so the better summary is Summary 2. In the second case, both summaries are factually consistent. However, Summary 1 is more comprehensive, so we prefer it to Summary 2. On the other hand, the annotators are supposed to choose the one with minor mistakes, while both summaries have hallucinations. In the last case, it is hard to tell which hallucination is more "acceptable". Considering that, we allow the workers to annotate it as "0" while it is hard to choose a better one.

\section{The human evaluation for coherence and content coverage.}
\label{sec:appendixh}

\begin{table*}[t!]
    \centering
    \resizebox{1.0\linewidth}{!}{
    \begin{tabular}{p{2.0\columnwidth}}\hline
         \textcolor{blue}{Article}: Even though the UK as a whole voted to leave the EU, in Scotland most people voted to remain. Now, some people are saying that Scotland should get independence from the rest of the UK, so that it can join the EU again on its own. Naz has been in Scotland to see what kids there think.\\
         \textcolor{blue}{Summary 1}: \textcolor{red}{Scottish children want independence from the UK so they can rejoin the EU}, despite the majority of the UK voting to leave.\\
         \textcolor{blue}{Summary 2}: Scotland voted to remain in the EU, but most of the UK voted to leave. Some Scots are now calling for independence to rejoin the EU.\\
         \textcolor{blue}{Which is better?}: \textbf{2}\\
         \hline
         \textcolor{blue}{Article}: Concerns had been raised by the Pakistan Cricket Board over threats from Hindu extremists, who attacked the offices of Indian cricket's governing body last year. However, PCB chairman Shaharyar Khan said on Thursday that the team had been cleared to play in March and April. "As a duty of care, we have asked [world cricket governing body] the ICC to put in place special arrangements for the Pakistan cricket team while in India." The World T20 runs from 8 March to 3 April...\\
         \textcolor{blue}{Summary 1}: Pakistan's cricket team has been cleared to tour India in March and April for the World T20, despite concerns over Hindu extremist threats. The PCB has asked the ICC to provide special security arrangements for the team.\\
         \textcolor{blue}{Summary 2}: The Pakistan cricket team has been given clearance to play in India in March and April for the World T20, despite concerns over threats from extremists.\\
         \textcolor{blue}{Which is better?}: \textbf{1}\\
         \hline
         \textcolor{blue}{Article}: Lambing season is a busy time for farmers, with thousands of baby sheep being born across the UK. Ten-year-old Tom and his sister Mali have been helping their family during lambing season. Around 4,000 lambs will be born on Tom and Mali's farm this year. Leah visits north Wales to meet them, and to find out just how busy it can be...\\
         \textcolor{blue}{Summary 1}: Farming siblings Tom and Mali have been lambing sheep for the \textcolor{red}{last three years}.\\
         \textcolor{blue}{Summary 2}: Tom and Mali, aged 10 and \textcolor{red}{12}, are helping their family on their farm in north Wales during Lambing season. They expect to birth around 4000 lams this year.\\
         \textcolor{blue}{Which is better?}: \textbf{0}\\
         \hline
    \end{tabular}
    }
    \caption{Three examples of human evaluation.}
    \label{tab:evalcases}
\end{table*}

\begin{table}[t!]
    \centering
    \resizebox{1.0\linewidth}{!}{
    \begin{tabular}{lcccc} \hline  
          \multicolumn{2}{c}{Models}&  Mean&  Max& Min \\ \hline  
          \multirow{2}{*}{LLaMA2 (7B)}&vanilla&  0.7005&  0.7545&  0.6502 \\
          &trained&  0.7045&  0.7722&  0.6598 \\
          \hline
          \multirow{2}{*}{Koala (7B)}&vanilla&  0.6898&  0.7421&  0.6283 \\
          &trained&  0.6951&  0.7572&  0.6364 \\
         \hline 
         \multirow{2}{*}{Tulu (7B)}&vanilla&  0.6871&  0.7613&  0.5953 \\
          &trained&  0.7013&  0.8011&  0.6310 \\
         \hline 
          \multirow{2}{*}{Vicuna (7B)}&vanilla&  0.6862&  0.7476&  0.6296 \\
          &trained&  0.7055&  0.7545&  0.6543 \\
         \hline 
    \end{tabular}
    }
    \caption{Statistics of head-level probing scores (the mean, maximum and minimum of probing scores of all the heads).}
    \label{tab:probingstatistics}
\end{table}

Figure~\ref{Fig:winc} and Figure~\ref{Fig:wincover} show the win rate of CPO+PST on coherence and content coverage, respectively. Coherence means logical, orderly, and consistent relation of parts, and content coverage indicates how many critical points of the original text are covered by summary. The human evaluation criterion is similar to Appendix~\ref{sec:appendixg}. The workers must annotate the index of the better one.

Compared with the factual consistency, the increase in coherence is slight. As for the content coverage, CPO+PST shows a noticeable improvement. The enhanced model tends to summarize in a more general way instead of a detailed description, which may be one reason for improving factual consistency and content coverage.

\section{More examples about decoupling.}
\label{sec:appendixb}

This section shows the difference between outputs under $I^C$ and $I^I$. As shown in Table~\ref{tab:cases}, given the same article, there are obvious differences between their generations. In the first example, $I^I$ writes what will happen as what has already happened. In the second example, $I^I$ adds a specific year, which does not appear in the source text. In the last example, $I^I$ confuses place names and replaces "Africans" with "African students". As mentioned in Section~\ref{sec:doesdecouple}, the hallucinations are not entirely irrelevant to the source text and seem like an adaptation of the original article. In other words, $I^I$ does not fabricate without any basis but embellishes the source text.

\section{The head-level probing scores.}
\label{sec:appendixc}

This section shows the mean, maximum and minimum of head-level probing scores. As shown in Table~\ref{tab:probingstatistics}, after being trained by CPO with PST, the attention heads of backbones achieve a higher probing score, which means they have a stronger ability to distinguish consistent and inconsistent summaries. The conclusion is aligned with the visualization in Section~\ref{sec:faccon}.

\end{document}